\pdfoutput=1
\documentclass[lettersize,journal]{IEEEtran}
\usepackage{amsmath,amsfonts}
\usepackage{algorithmic}
\usepackage{algorithm}
\usepackage{array}
\usepackage[caption=false,font=normalsize,labelfont=sf,textfont=sf]{subfig}
\usepackage{textcomp}
\usepackage{stfloats}
\usepackage{url}
\usepackage{verbatim}
\usepackage{graphicx}
\usepackage{epstopdf}
\usepackage{multirow}
\usepackage{cite}
\usepackage{pifont}
\usepackage{booktabs}

\begin{document}

\title{Unbiased Scene Graph Generation by\\ Type-Aware Message Passing \\on Heterogeneous and Dual Graphs}

\author{Guanglu Sun, Jin Qiu, Lili Liang\\
	Harbin University of Science and Technology, School of Computer Science and Technology\\ Harbin, 157000\\
	
\thanks{This paper was produced by the IEEE Publication Technology Group. They are in Piscataway, NJ.}
\thanks{Manuscript received April 19, 2021; revised August 16, 2021.}}

\markboth{Journal of \LaTeX\ Class Files,~Vol.~14, No.~8, August~2021}%
{Shell \MakeLowercase{\textit{et al.}}: A Sample Article Using IEEEtran.cls for IEEE Journals}


\maketitle

\begin{abstract}
	Although great progress has been made in the research of unbiased scene graph generation, issues still hinder improving the predictive performance of both head and tail classes. An unbiased scene graph generation (TA-HDG) is proposed to address these issues. For modeling interactive and non-interactive relations, the Interactive Graph Construction is proposed to model the dependence of relations on objects by combining heterogeneous and dual graph, when modeling relations between multiple objects. It also implements a subject-object pair selection strategy to reduce meaningless edges. Moreover, the Type-Aware Message Passing enhances the understanding of complex interactions by capturing intra- and inter-type context in the Intra-Type and Inter-Type stages. The Intra-Type stage captures the semantic context of inter-relaitons and inter-objects. On this basis, the Inter-Type stage captures the context between objects and relations for interactive and non-interactive relations, respectively. Experiments on two datasets show that TA-HDG achieves improvements in the metrics of R@K and mR@K, which proves that TA-HDG can accurately predict the tail class while maintaining the competitive performance of the head class.
\end{abstract}

\begin{IEEEkeywords}
Scene graph generation, message passing, graph construction.
\end{IEEEkeywords}

\section{Introduction}
\IEEEPARstart{S}{cene} Graph Generation (SGG) aims to identify objects and their relations in images. Because of its potential applications in image generation \cite{liu2024r3cd}, visual question answering \cite{qian2022scene}, image captioning \cite{wang2021high}, etc., SGG has become a hot topic in the field of computer vision. Currently, SGG methods \cite{zheng2023prototype,jiang2023scene} follow a conventional two-stage strategy \cite{li2024scene}, which involves first detecting objects and then predicting the relationships between them. Although these methods have made some progress, they tend to favor the head classes owing to the unbalanced distribution of relations. Therefore, a significant challenge in SGG is accurately predicting tail classes while maintaining competitive performance on head classes.

\begin{figure}[htbp]
	\centering
	\setlength{\belowcaptionskip}{-0.5cm}
	\includegraphics[width=83mm]{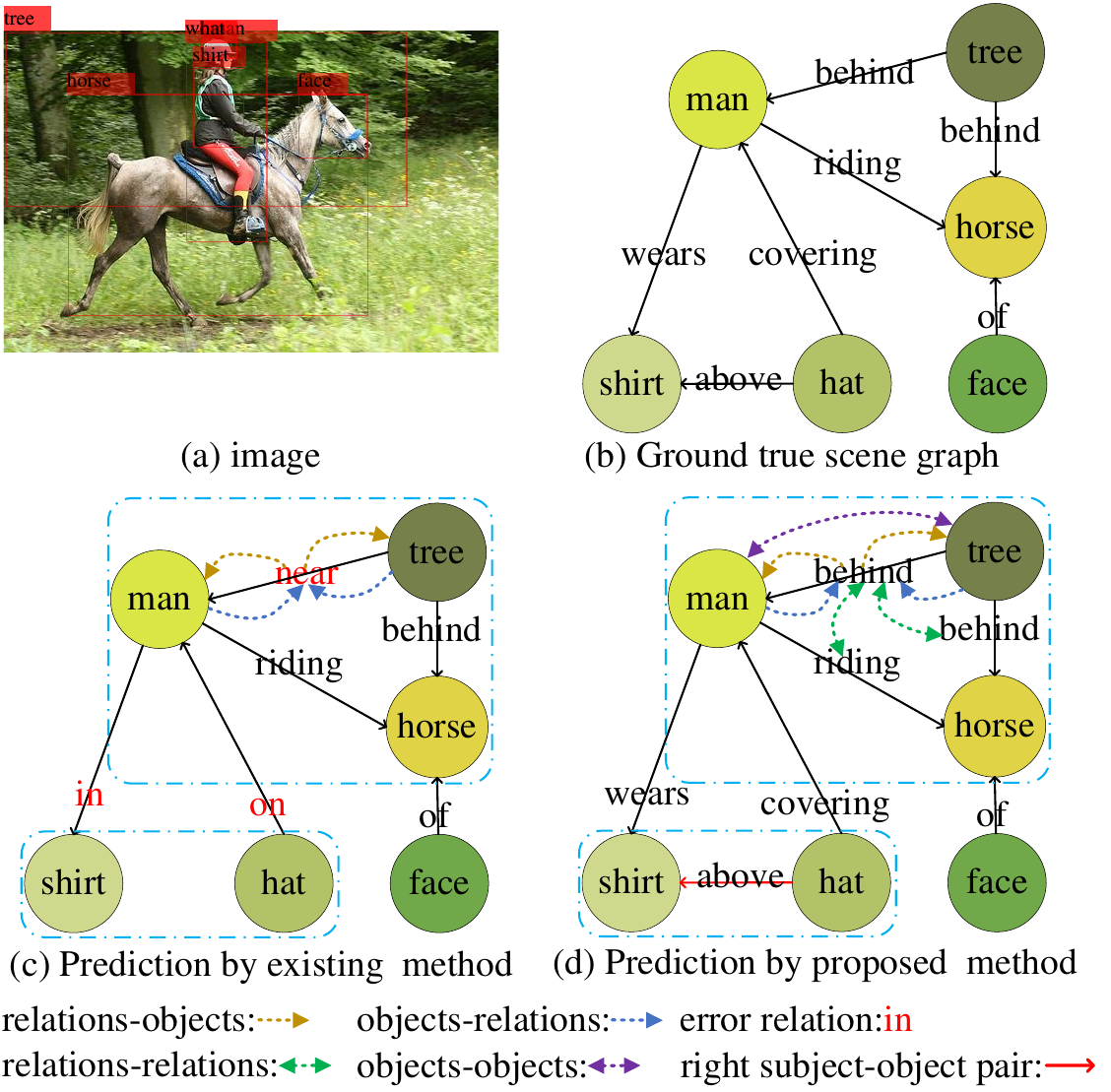}
	\caption{(a) Input image. (b) Ground true scene graph. (c) Prediction by existing SGG method. (d) Prediction by proposed SGG method. The round boxes represent objects, solid lines represent relations, dashed lines represent the direction of message passing, the red solid lines represent right subject-object pairs, and red words represent prediction errors.
	}
	\label{introduction}	
\end{figure}
Existing unbiased methods construct graphs to model relations, then pass messages on the graphs to capture context, thereby improving the predictive performance on tail classes \cite{latipova2023overview}. Specifically, in terms of modeling relations, HetSGG \cite{yoon2023unbiased} constructs an object-centric heterogeneous graph to model the dependence of relations on objects. EdgeSGG \cite{kim2023semantic} builds a relation-centric dual graph to model relations between multiple objects. Some methods \cite{zhou2022unified,tian2020part} select high-confidence detected objects to construct sparse graphs. However, these methods ignore the interactions between relations when modeling the interactions between objects, leading to the inability to model interactions between multiple relations. Here, these interactions are categorized as intra-type interactions, encompassing two types——objects and relations. As shown in Fig.~\ref{introduction}(c), due to the existing methods do not model the interaction between the relations between \textit{hat-man} and \textit{man-shirt}, the relation between \textit{hat} and \textit{shirt} is not predicted. Therefore, modeling both the interactions between relations and the interactions between objects enables models to construct an effective graph structure to comprehensively represent the relations of objects in the scene.

Common methods aggregate context between objects and relations by message passing. Specifically, some methods \cite{lin2020gps,li2021bipartite} combine the direction information of relations to aggregate direction-aware context via the graph neural network. Some methods \cite{yoon2023unbiased,kim2023semantic} use the attention mechanism to adaptively adjust the aggregation weight and aggregate context from neighboring objects to refine the features of objects and relations. However, these methods only focus on the context between objects and relations (i.e., inter-type context), and ignore capturing semantic context among different objects with the same relation and among different relations with the same object (i.e., intra-type context). The neglect of intra-type context limits these models to precisely refine the features of objects and relations, thereby reducing the accuracy in relation prediction. Fig.~\ref{introduction}(c) shows that the existing methods do not capture the intra-type context of relations-relations, such as the \textit{man} \textit{riding} the \textit{horse} and the \textit{horse} \textit{behind} the \textit{tree}, leading to no inference that the \textit{man} is also \textit{behind} the \textit{tree}. Therefore, capturing intra-type context and refining features helps to fully understand and represent complex interactions within a scene, improving the accurate prediction of relations.

To address these issues, an Unbiased Scene Graph Generation by Type-Aware Message Passing on Heterogeneous and Dual Graphs (TA-HDG) is designed to improve the performance of head and tail classes. It categorizes relation classes into interactive and non-interactive relation types to adjust the distribution of relations. To model interactions and capture context for these relation types, Heterogeneous and Dual Graph Construction (HDGC) is proposed to model the intra-type interactions by combining heterogeneous and dual graphs. To reduce meaningless edges in the graphs and enhance the precision of subject-object pairs, HDGC utilizes a subject-object pair selection strategy, which introduces extra information. The information contains distance, confidence, and existence information. Moreover, Type-Aware Message Passing (TAMP) is proposed to precisely refine the features of objects and relations and enhance the understanding of complex interactions by capturing intra- and inter-type context in the Intra-Type and Inter-Type stages. In the Intra-Type stage, messages of interactions between objects and interactions between relations (i.e., intra-type interactions) are passed on the dual graph to capture the intra-type context. In the Inter-Type stage, for interactive and non-interactive relations, messages of interactions between objects and relations (i.e., inter-type interactions) are passed on the heterogeneous graph to capture the inter-type context. Fig.~\ref{introduction}(d) shows that TA-HDG successfully identifies the \textit{hat-shirt} pair by modeling the intra-type interactions, and it correctly predicts the relation of \textit{man-tree} by capturing the intra- and inter-type context via the inter-type message passing (relations-objects and objects-relations), as well as the intra-type message passing (relations-relations and objects-objects).

The contributions of this study are as follows: 
\begin{itemize}
	\item{TA-HDG is an unbiased SGG framework that alleviates the long-tail problem by categorizing the relation classes into balanced interactive and non-interactive relation types, and modeling and aggregating their contextual information. TA-HDG achieves state-of-the-art performance on Visual Genome and Open Images datasets.}
	\item{The Heterogeneous and Dual Graph Construction models the interactions between objects and interactions between relations. It introduces distance, confidence, and existence information to reduce meaningless edges in the graphs.}
	\item{The Type-Aware Message Passing captures intra- and inter-type context by achieving the Intra-Type message passing and the Inter-Type message passing to enhance the understanding of complex interactions for improving the accurate prediction of relations.}	
\end{itemize}

\section{Related Work}
\subsection{Unbiased Scene Graph Generation}
Although the SGG task has made progress, the long-tail distribution of relations leads models to favor the general relations. This limits the application of the SGG task in other fields. Recent studies \cite{zhao2021semantically,zheng2023dual,li2023label} focus on unbiased scene graph generation. BGNN and PSCV \cite{li2021bipartite,zhou2022peer} utilize the re-sampling strategy to enrich the samples. EOA \cite{chen2023more} introduces external knowledge for data enhancement. TGDA TGDA \cite{zang2024template} generates relation templates based on knowledge distillation to provide supplementary training data. The data enhancement techniques designed by these methods are for the whole dataset and may not address the issue of long-tail distribution. Therefore, some methods \cite{liu2023importance,liu2023constrained} introduce the re-weighting strategy to adjust weights for each relation, making models favor the tail classes. However, these studies neglect to maintain competitive performance on head classes.

In this work, TA-HDG improves the performance on head and tail classes by modeling and capturing the interactions for the interactive and non-interactive relation types, respectively.

\subsection{Graph Construction in Scene Graph Generation}
In the SGG task, constructing an effective graph can better model the relations among objects. IMP\cite{xu2017scene} models the relations by constructing a fully connected graph, mistakenly selecting subject-object pairs. Therefore, some methods \cite{zareian2020bridging,chen2023more,wang2023novel} introduce external knowledge to build sparse graphs, removing some meaningless edges. The mismatch between external knowledge and SGG datasets still leads to incorrect selection. To address the limitations of these methods, some methods\cite{zhou2022unified, tian2020part,lin2022ru,liu2023neural} filter irrelevant subject-object pairs by introducing the confidence scores of detected objects or the structural Bethe approximation. On the other hand, HetSGG \cite{yoon2023unbiased} constructs a heterogeneous graph to model the dependence of relations on objects. EdgeSGG \cite{kim2023semantic} builds a dual graph to model relations between multiple objects. However, these methods ignore the interactions between relations when modeling the interactions between objects, leading to the inability to model interactions between multiple relations.

In this work, HDGC combines heterogeneous and dual graphs to model both the interactions between objects and the interactions between relations. And it designs a subject-object pair selection strategy to reduce meaningless edges.

\subsection{Massage Passing in Scene Graph Generation}
To improve the prediction of objects and relations in SGG, some studies \cite{xu2017scene,zellers2018neural} capture the context around objects through the message passing to refine their features. For example, some works \cite{lin2020gps,li2021bipartite} combine the direction information of relations to aggregate direction-aware context via the graph neural network. Some methods \cite{yoon2023unbiased,kim2023semantic,yang2024adaptive} use the attention mechanism to adaptively adjust the aggregation weight and aggregate context from neighboring objects to refine the features of objects and relations. Some studies \cite{zhong2021learning,chiou2021recovering} pass messages between visual features and textual embedding by Transformer or LSTM. However, these methods only focus on the context between objects and relations (that is, inter-type message passing between objects and relations) and ignore the semantic context among different objects with the same relation and among different relations with the same object (that is, intra-type message passing of objects and intra-type message passing of relations). This limits these models to fully understanding and representing complex interactions within a scene, thereby reducing the accuracy in relation prediction. 

Unlike previous methods, TAMP captures the intra-type context and the inter-type context, thereby enhancing the understanding of the scene and improving the accurate prediction of relations.

\begin{figure*}[htbp]
	\centering
	\setlength{\belowcaptionskip}{-0.3cm}
	\includegraphics[width=180mm]{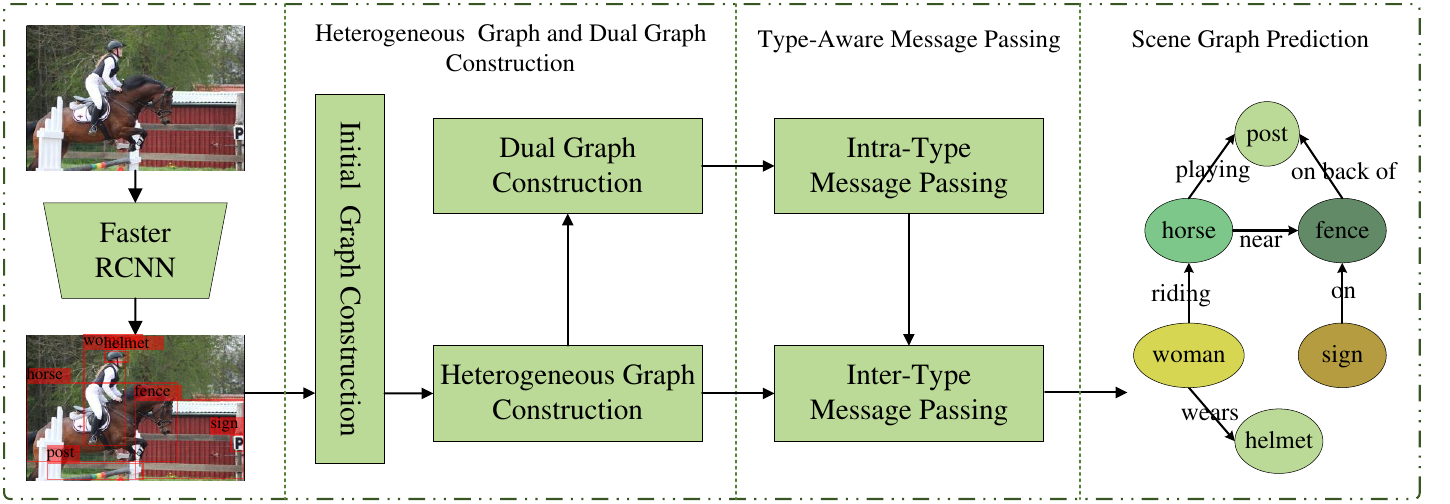}
	\caption{The architecture of TA-HDG. First, object proposals are obtained by Faster R-CNN. Second, HDGC introduces object information (distance, confidence, and existence information) to select subject-object pairs for constructing an initial graph. Then it constructs a heterogeneous graph to model the interactions between objects and constructs a dual graph to model the interactions between relations. Third, TAMP refines the features of objects and relations via the Intra-Type Message Passing on the dual graph and the Inter-Type Message Passing on the heterogeneous graph. Finally, a scene graph is generated based on the features of objects and relations.}
	\label{method}
\end{figure*}

\section{Method}
Fig.~\ref{method} illustrates the architecture of TA-HDG. The subsequent sections describe in detail the model. First, HDGC introduces object information to select subject-object pairs, then it not only constructs a heterogeneous graph to model the interactions between objects but also constructs a dual graph to model the interactions between relations (Sec.~\ref{3.1}). Second, TAMP captures the context of intra- and inter-type interactions via the Intra-Type message passing and the Inter-Type message passing, respectively (Sec.~\ref{3.2}). Finally, the details of scene graph prediction and model training are presented (Sec.~\ref{3.3}).

Before presenting the method, the problem definition is introduced first. For a given image I, the goal is to generate a scene graph $G= \left \langle {V,E} \right \rangle$, where $V$ is the set of detected objects, $E$ is the set of relations of subject-object pairs. Each object $o_i \in {V}$ is composed of the bounding box $b_i$ and the class label $c_i \in \mathbf{C}$, where $\mathbf{C}$ is the set of object classes. Each relation $r_{i \rightarrow j}\in {E}$ between subject $o_i \in {V}$ and object $o_j \in {V}$ has its corresponding relation label in the set of relation classes $\mathbf{R}$.

\subsection{Heterogeneous and Dual Graph Construction}\label{3.1}
\textbf{Initial Graph Construction.} First, the Faster R-CNN \cite{ren2016faster} is utilized to obtain the information of proposal objects, which includes the bounding box ${b}_i$, the visual feature $v_i$, the class label $c_i$, and the class distribution $p_i$. Then, these objects are connected pairwise to construct a fully connected graph $G^\mathrm{o}= \left \langle {V,E} \right \rangle$ with detected objects as nodes and relations between subject-object pairs as edges. For the object $o_i$ in $G^\mathrm{o}$, its object feature $f_i$ can be represented as:

\begin{small}
	\begin{equation}
		f_i \!=\! \mathbf{W}_\mathrm{o}\left[\mathbf{W}_\mathrm{v} {v}_i; \mathbf{W}_\mathrm{b} {b}_i; {c}_i\right] \label{con:eq_2}
	\end{equation}
\end{small}where $\mathbf{W}_\mathrm{o}$, $\mathbf{W}_\mathrm{v}$, and $\mathbf{W}_\mathrm{b}$ are linear transformation matrices, $\left[ \cdot;\cdot;\cdot\right]$ is the concatenation operation. For the relation $r_{i \rightarrow j}$ between $o_i$ and $o_j$ in $G^\mathrm{o}$, its relation feature $f_{i \rightarrow j}$ can be represented as:

\begin{small}
	\begin{equation}
		f_{i \rightarrow j} \!=\! \mathbf{W}_\mathrm{r}\left[f_i; f_j; b_{i \rightarrow j}\right] \label{con:eq_3}
	\end{equation}
\end{small}where $\mathbf{W}_\mathrm{r}$ is a linear transformation matrix, $b_{i \rightarrow j}$ is the union bounding box of $o_i$and $o_j$. 

To reduce meaningless edges, the subject-object pair selection strategy is designed, which utilizes object information (including distance, confidence, and existence information). Specifically, the distance matrix of subject-object pairs $\mathbf{M}_\mathrm{b}$ calculated based on the bounding boxes measures the spatial information between subject-object pairs, the confidence matrix of subject-object pairs $\mathbf{M}_\mathrm{p}$ calculated based on the class distributions measures the confidence information between subject-object pairs, the existence matrix of subject-object pairs $\mathbf{M}_\mathrm{l}$ calculated based on the class labels and co-linearities measures the semantic information between subject-object pairs. Finally, these matrices filtered by three thresholds are applied to retain meaningful subject-object pairs: $s_\mathrm{b}$, $s_\mathrm{l}$, and $s_\mathrm{p}$:

\begin{small}
	\begin{equation}		
		\hat{E} \!=\! \left(\mathbf{M}_\mathrm{b} <s_\mathrm{b}\right) \cap\left(\mathbf{M}_\mathrm{l} >s_\mathrm{l}\right) \cap\left(\mathbf{M}_\mathrm{p} > s_\mathrm{p}\right) \label{con:eq_1}
	\end{equation}
\end{small}where $s_\mathrm{b}$ and $s_\mathrm{l}$ are hyperparameters, $s_\mathrm{p}$ is the confidence score of the $top \mbox{-}K \mbox{-}\mathrm{th}$ element from ranked $\mathbf{M}_\mathrm{p}$. The initial graph $\hat{G}^\mathrm{o}= \left \langle {V,\hat{E}} \right \rangle$ can be then obtained. The nodes are detected objects, and the edges are the retained meaningful relations.

\textbf{Heterogeneous Graph Construction.} To adjust the distribution of each relation, the relation super-types in Motifs \cite{zellers2018neural} are categorized into two relation types: interactive relations and non-interactive relations. Subsequently, to model the interactions between objects for these relation types, inspired by HetSGG \cite{yoon2023unbiased}, the initial graph $\hat{G}^\mathrm{o}= \left \langle {V,\hat{E}} \right \rangle$ is converted into a heterogeneous graph $G^\mathrm{h} \!=\! \left \langle {V,E^\mathrm{h}} \right \rangle$. For the acquisition process of $E^\mathrm{h}$, the relation feature $f_{i \rightarrow j}$ is pre-classified by a linear classifier: $p_{i \rightarrow j}^\mathrm{o} \!=\! \operatorname{softmax}\left(\mathbf{W}_\mathrm{rel}^\mathrm{o} f_{i \rightarrow j}\right)$. Each element of $p_{i \rightarrow j}^\mathrm{o}$ represents the probability for a specific relation class. The relation type $t_{i \rightarrow j}$ is then inferred via a pre-defined function $\gamma$, which maps the relation classes to the relation types, $\gamma: \mathbf{C} \rightarrow \mathbf{T}$. $\mathbf{T}$ includes interactive relation type ($\varphi$) and non-interactive relation type ($\delta$). Note that the $\mathbf{Mean \left( \cdot \right)}$ is utilized as the pre-defined function $\gamma$. Finally, a heterogeneous graph is constructed with objects as nodes and two types of relations as edges.

\textbf{Dual Graph Construction.} A dual graph is introduced to model the interactions between relations. The dual graph $G^\mathrm{d}\!=\!\left \langle{V^\mathrm{d},E^\mathrm{d}}\right \rangle$ is constructed by exchanging nodes and edges in the heterogeneous graph $G^\mathrm{h} \!=\! \left \langle {V,E^\mathrm{h}} \right \rangle$. For the acquisition process of $V^\mathrm{d}$, first, the instances of visual phrases in the heterogeneous graph are counted, i.e. $\left \langle \textit{subject-relation-object} \right \rangle$. Subsequently, for each visual phrase, the relation is taken as the node in $G^\mathrm{d}$, i.e. $V^\mathrm{d} \!=\! \left\{r_{i \rightarrow j}^\mathrm{d} \mid r_{i \rightarrow j}^\mathrm{d} \in E^\mathrm{h}\right\}$. For the acquisition process of $E^\mathrm{d}$, for relations in any two visual phrases, estimate whether they are involved in the same subject or object. If there is the same subject or object, an edge is formed by the shared subject or object between relations, i.e. $E^\mathrm{d} \!=\! \left\{o_{i}^\mathrm{d} \!=\! \left(r_{j \rightarrow i}^\mathrm{h}, r_{i \rightarrow k}^\mathrm{h}\right) \mid r_{j \rightarrow i}^\mathrm{h} \cap r_{i \rightarrow k}^\mathrm{h} \!=\! o_{i}^\mathrm{d} \in V\right\}$. Eventually, a dual graph is constructed with relations as nodes and objects as edges.

\subsection{Type-Aware Message Passing}\label{3.2}
\textbf{Intra-Type Message Passing (Intra-MP.)} As shown in Fig.~\ref{intra}, Intra-MP refines the object and relation features via relations-relations and objects-objects on the dual graph to capture the messages of intra-type interactions. Thus, the semantic context among different relations with the same object and among different objects with the same relation is learned. It consists of the following two steps:

\textbf{1)update of relations-relations.} For the relation $r_{i \rightarrow j}^\mathrm{d}$ in the dual graph, the attention mechanism is utilized to update the relation feature by passing neighboring information of the relation to the relation. This process allows the model to learn the semantic context among different relations with the same object. Specifically, the neighboring information of the relation $r_{i \rightarrow j}^\mathrm{d}$ refers to the relation features of all other relations $r_{i \rightarrow k}^\mathrm{d}$, which are directly connected to the relation $r_{i \rightarrow j}^\mathrm{d}$. The formula for updating relation features is as follows:

\begin{figure}[htbp]
	\centering
	\setlength{\belowcaptionskip}{-0.3cm}
	\includegraphics[width=87mm]{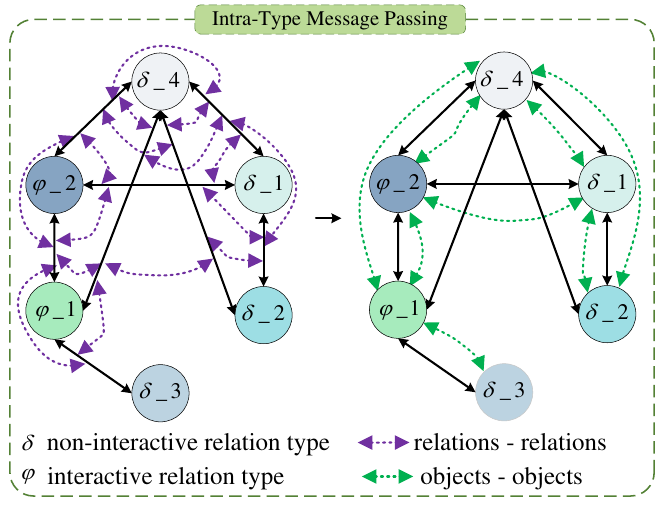}
	\caption{The Intra-Type Message Passing refines features via relations-relations and objects-objects on the dual graph.}
	\label{intra}
\end{figure}
\begin{small}
	\begin{equation}
		f_{i \rightarrow j}^{\mathrm{d},(l+1)} \!=\! f_{i \rightarrow j}^{\mathrm{d},(l)} +\psi\left(\sum_{r_{i \rightarrow k}^\mathrm{d} \in \mathcal{N}_{r_{i \rightarrow j}^\mathrm{d}}} \alpha_{ij \rightarrow ik}^{\mathrm{d},(l)}
		\mathbf{W}_\mathrm{r}^\mathrm{d} f_{i \rightarrow k}^{\mathrm{d},(l)}\right) \label{con:eq_4}
	\end{equation}
\end{small}where $f_{i \rightarrow j}^{\mathrm{d},(l+1)}$ is the relation feature of $r_{i \rightarrow j}^\mathrm{d}$ at the $(l+1)\mbox{-} \mathrm{th}$ layer. The initial representation of $f_{i \rightarrow j}^{\mathrm{d},(l+1)}$ is the relation feature in the initial graph $\hat{G}^\mathrm{o}$: $f_{i \rightarrow j}^{\mathrm{d},(0)} \!=\! f_{i \rightarrow j}$. $\psi$ is an activation function (e.g. ReLU). $\mathcal{N}_{r_{i \rightarrow j}^\mathrm{d}}$ is the set of relations, which are the neighbors of relation $r_{i \rightarrow j}^\mathrm{d}$. $\mathbf{W}_\mathrm{r}^\mathrm{d}$ is the weight matrix. $\alpha_{ij \rightarrow ik}^{\mathrm{d},(l)}$ is the attention score calculated via the weight matrix $\mathbf{W}_\mathrm{att,d}$ as follows:

\begin{small}
	\begin{equation}
		\alpha_{ij \rightarrow ik}^{\mathrm{d},(l)} \!=\! \frac{\exp \left( \mathbf{W}_\mathrm{att,d}^\mathrm{T} \left[ f_{i \rightarrow j}^{\mathrm{d},(l)}; f_{i \rightarrow k}^{\mathrm{d},(l)} \right] \right)} {\sum_{r_{i \rightarrow q}^\mathrm{d} \in \mathcal{N}_{r_{i \rightarrow j}^\mathrm{d}}} \exp \left( \mathbf{W}_\mathrm{att,d}^\mathrm{T} \left[ f_{i \rightarrow j}^{\mathrm{d},(l)} ; f_{i \rightarrow q}^{\mathrm{d},(l)} \right] \right)} \label{con:eq_5}
	\end{equation}
\end{small}where $\left[ \cdot;\cdot\right]$ is the concatenation operation.

\textbf{2)update of objects-objects.} For the object $o_{i}^\mathrm{d}$ in the dual graph, the attention mechanism is utilized to update the object feature by passing neighboring information of the object to the object. This process allows the model to learn the semantic context among different objects with the same relation. Specifically, the neighboring information of the object $o_{i}^\mathrm{d}$ refers to the object features of all other objects $o_{j}^\mathrm{d}$, which are directly connected to the object $e_{ij}^\mathrm{d}$. The formula for updating object features is as follows:

\begin{small}
	\begin{equation}
		f_i^{\mathrm{d},(l+1)}=f_i^{\mathrm{d},(l)}+\psi\left(\sum_{o_j^\mathrm{d} \in \mathcal{N}_{o_i^\mathrm{d}}} \alpha_{i \rightarrow j}^{\mathrm{d},(l)}
		\mathbf{W}_\mathrm{o}^\mathrm{d} f_j^{\mathrm{d},(l)}\right) \label{con:eq_6}
	\end{equation}
\end{small}where $f_i^{\mathrm{d},(l+1)}$ is the object feature of $o_{i}^\mathrm{d}$ at the $(l+1)\mbox{-} \mathrm{th}$ layer. The initial representation of $f_i^{\mathrm{d},(l+1)}$ is the object feature in the initial graph $\hat{G}^\mathrm{o}$: $f_i^{\mathrm{d},(0)} \!=\! f_{i}$. $\mathcal{N}_{o_i^\mathrm{d}}$ is the set of objects, which are the neighbors of object $o_i^\mathrm{d}$. $\mathbf{W}_\mathrm{o}^\mathrm{d}$ is the weight matrix. $\alpha_{i \rightarrow j}^{\mathrm{d},(l)}$ is the attention score calculated in the same way following Eq.~\ref{con:eq_6}.

\begin{figure}[htbp]
	\centering
	\setlength{\belowcaptionskip}{-0.3cm}
	\includegraphics[width=87mm]{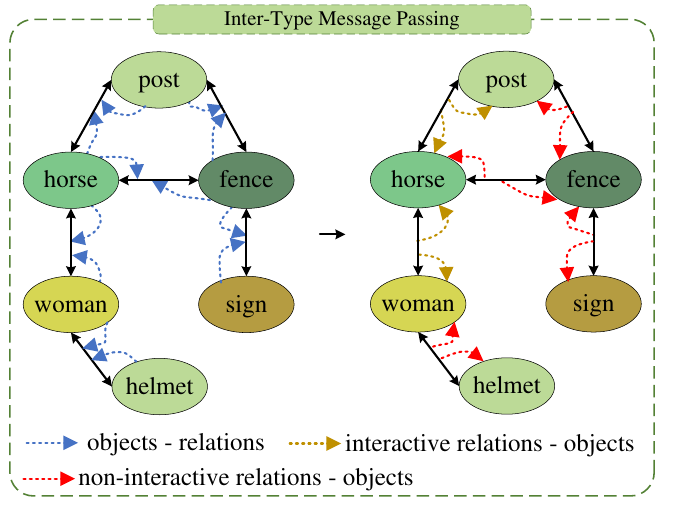}
	\caption{The process of Inter-Type Message Passing refines features via objects-relations and relations-objects on the heterogeneous graph.}
	\label{inter}
\end{figure}
\textbf{Inter-Type Message Passing (Inter-MP).} As shown in Fig.~\ref{inter}, Inter-MP refines the object and relation features via objects-relations and relations-objects on the heterogeneous graph to capture the messages of inter-type interactions. Thus, the semantic context between objects and relations is learned. Here, the messages of inter-type interactions are involved in the interactive contextual information, as well as the non-interactive contextual information. It consists of the following two steps:

\textbf{1)update of objects-relations.} For the relation $r_{i \rightarrow j}^{\mathrm{h}}$ in the heterogeneous graph with the relation type $t_{i \rightarrow j}$, the attention mechanism is utilized to update the relation feature by passing neighboring information of the relation to the relation. This process allows the model to learn the dependence of relations on objects. Specifically, the neighboring information of the relation $r_{i \rightarrow j}^{\mathrm{h}}$ refers to the two objects $o_i^{\mathrm{h}}$ and $o_j^{\mathrm{h}}$, which are directly connected to the relation $r_{i \rightarrow j}^{\mathrm{h}}$. The $o_i^{\mathrm{h}}$ and $o_j^{\mathrm{h}}$ are served as the subject and object for the relation $r_{i \rightarrow j}^{\mathrm{h}}$ in a visual phrase, respectively. The formula for updating relation features is as follows:

\begin{small}
	\begin{equation}
		\begin{aligned}
			f_{i \rightarrow j}^{\mathrm{h},(l+1)} \!=\! f_{i \rightarrow j}^{\mathrm{h},(l)} +
			\psi \left[ 
			{\alpha_{i \rightarrow j}^{\mathrm{h},(l)} \mathbf{W}_{t_{i \rightarrow j}}^i f_i^{\mathrm{h},(l)} } \right. \\
			+  
			\left. {\left( 1-\alpha_{i \rightarrow j}^{\mathrm{h},(l)}\right)
				\mathbf{W}_{t_{i \rightarrow j}}^j f_j^{\mathrm{h},(l)}} \right] \\	
		\end{aligned}	
		\label{con:eq_8}
	\end{equation}
\end{small}where $\mathbf{W}_{t_{i \rightarrow j}}^i$ and $\mathbf{W}_{t_{i \rightarrow j}}^j$ are the weight matrices of the relation type $t_{i \rightarrow j}$, representing the messages passed from object $o_i^{\mathrm{h}}$ to relation $r_{i \rightarrow j}^{\mathrm{h}}$ and from object $o_j^{\mathrm{h}}$ to relation $r_{i \rightarrow j}^{\mathrm{h}}$, respectively. $f_{i \rightarrow j}^{\mathrm{h},(l+1)}$ is the relation feature of $r_{i \rightarrow j}^{\mathrm{h}}$ at the $(l+1) \mbox{-} \mathrm{th}$ layer, $f_i^{\mathrm{h},(l)}$ and $f_j^{\mathrm{h},(l)}$ are the object features of $o_i^{\mathrm{h}}$ and $o_j^{\mathrm{h}}$ at the $(l+1) \mbox{-} \mathrm{th}$ layer, respectively. The initial representations of objects and relations are the refined features in the dual graph $G^\mathrm{d}$: $f_{i \rightarrow j}^{\mathrm{h},(0)} \!=\! f_{i \rightarrow j}^{\mathrm{d},(l+1)}$,$f_{i}^{\mathrm{h},(0)} \!=\! f_{i}^{\mathrm{d},(l+1)}$. $\alpha_{i \rightarrow j}^{\mathrm{h},(l)}$ is the attention score calculated via the weight matrix $\mathbf{W}_\mathrm{att,h}$ as follows:

\begin{small}
	\begin{equation}
		\alpha_{i \rightarrow j}^{\mathrm{h},(l)} \!=\! \frac{\exp \left(\mathbf{W}_\mathrm{att,h}^\mathrm{T} f_i^{\mathrm{h},(l)}\right)} {\exp \left(\mathbf{W}_\mathrm{att,h}^\mathrm{T} f_i^{\mathrm{h},(l)}\right)+\exp \left(\mathbf{W}_\mathrm{att,h}^\mathrm{T} f_j^{\mathrm{h},(l)}\right)}
		\label{con:eq_9}
	\end{equation}
\end{small}

\textbf{2)update of relations-objects.} According to the relation feature $f_{i \rightarrow j}^{\mathrm{h},(l+1)}$ obtained in the previous step, the object feature $f_i^{\mathrm{h}}$ is updated. The main idea is to aggregate information from neighboring relations with the same relation type. For the object $o_i^{\mathrm{h}}$ and the specified relation type $t_{i \rightarrow j}$ in the heterogeneous graph, the attention mechanism is utilized to update the object feature by passing neighboring information of the object to the object. This process allows the model to learn the dependence of objects on relations. Specifically, the neighboring information of the object $o_i^{\mathrm{h}}$ refers to all relations $r_{i \rightarrow j}^{\mathrm{h}}$ and $r_{j \rightarrow i}^{\mathrm{h}}$, which are directly connected to the object $o_i^{\mathrm{h}}$, and the relation type of these relations is $t_{i \rightarrow j}$. The $r_{i \rightarrow j}^{\mathrm{h}}$ and $r_{j \rightarrow i}^{\mathrm{h}}$ take the object $o_i^{\mathrm{h}}$ as the subject and object of a visual phrase, respectively. The formula for updating object features is as follows:

\begin{small}
	\begin{equation}
		\begin{aligned}
			f_{i, t_{i \rightarrow j}}^{\mathrm{h},(l+1)} \!=\! \sum_{r_{i \rightarrow j}^{\mathrm{h}} \in \mathcal{N}_{t_{i \rightarrow j}}}
			\left(
			{\alpha_{i \rightarrow j}^{t_{i \rightarrow j},\mathrm{\mathrm{h}},(l)} \mathbf{W}_{t_{i \rightarrow j}}^{i \rightarrow j} f_{i \rightarrow j}^{\mathrm{h},(l)}} \right. \\
			+
			\left. {\alpha_{j \rightarrow i}^{t_{i \rightarrow j},\mathrm{h},(l)} \mathbf{W}_{t_{i \rightarrow j}}^{j \rightarrow i} f_{j \rightarrow i}^{\mathrm{h},(l)} } \right)\\			
		\end{aligned}
		\label{con:eq_10}
	\end{equation}
\end{small}where $\mathbf{W}_{t_{i \rightarrow j}}^{i \rightarrow j}$ and $\mathbf{W}_{t_{i \rightarrow j}}^{j \rightarrow i}$ are the weight matrices of the relation type $t_{i \rightarrow j}$, representing the messages passed from relation $r_{i \rightarrow j}^{\mathrm{h}}$ to object $o_i^{\mathrm{h}}$ and from relation $r_{i \rightarrow j}^{\mathrm{h}}$ to object $o_j^{\mathrm{h}}$, respectively. $\mathcal{N}_{t_{i \rightarrow j}}$ is the set of relations with relation type $t_{i \rightarrow j}$, which are the neighbors of object $o_i^{\mathrm{h}}$. $\alpha_{i \rightarrow j}^{t_{i \rightarrow j},\mathrm{\mathrm{h}},(l)}$ is the attention score calculated via the weight matrix $\mathbf{W}_\mathrm{att,t}$ as follows:

\begin{small}
	\begin{equation}
		\alpha_{i \rightarrow j}^{t_{i \rightarrow j},\mathrm{\mathrm{h}},(l)}=\frac{\exp \left( \mathbf{W}_\mathrm{att,t}^\mathrm{T} f_{i \rightarrow j}^{\mathrm{h},(l)} \right) } {\sum_{r_{i \rightarrow k}^{\mathrm{h}} \in \mathcal{N}_{t_{i \rightarrow j}}} \exp \left( \mathbf{W}_\mathrm{att,t}^\mathrm{T} f_{i \rightarrow k}^{\mathrm{h},(l)} \right)}
		\label{con:eq_11}
	\end{equation}
\end{small}

Finally, to obtain the final object feature $f_i^{\mathrm{h},(l+1)}$, all the object features based on specific relation types are aggregated:

\begin{small}
	\begin{equation}
		f_i^{\mathrm{h},(l+1)}=f_i^{\mathrm{h},(l)}+\frac{1}{|\mathbf{T}|}\sum_{t_{i \rightarrow j}=1}^{|\mathbf{T}|} \psi \left( f_{i, t_{i \rightarrow j}}^{\mathrm{h},(l+1)} \right)
		\label{con:eq_12}
	\end{equation}
\end{small}where $\mathbf{T}$ is a set of relation types.

\subsection{Scene Graph Prediction}\label{3.3}
After obtaining the object features and relation features of the given image $I$, the class labels of objects and relations are predicted to generate a scene graph for this image. To be specific, for the obtained object feature $f_i^{\mathrm{h},(l+1)}$ and relation feature $f_{i \rightarrow j}^{\mathrm{h},(l+1)}$, two linear classifiers are applied to calculate the object class distribution $p_i$ and the relation class distribution $p_{i \rightarrow j}$, respectively:

\begin{small}
	\begin{equation}
		p_i \!=\! \operatorname{softmax}\left(\mathbf{W}_\mathrm{obj} f_i^{\mathrm{h},(l+1)}\right) \label{con:eq_13}
	\end{equation}
\end{small}
\begin{small}
	\begin{equation}
		p_{i  \rightarrow j} \!=\! \operatorname{softmax}\left( \mathbf{W}_\mathrm{rel} f_{i \rightarrow  j}^{\mathrm{h},(l+1)}\right) \label{con:eq_14}
	\end{equation}
\end{small}where $\mathbf{W}_\mathrm{obj}$ and $\mathbf{W}_\mathrm{rel}$ represent the weight matrices for the object and relation classifiers, respectively. Two loss functions are utilized to train TA-HDG, that is, binary cross entropy loss (BCE) \cite{ruby2020binary} for the object classification and binary cross entropy loss (BCE) \cite{ruby2020binary} for the relation classification. The two losses are added for the total loss:

\begin{small}
	\begin{equation}
		\mathcal{L} \!=\! \mathcal{L}_\mathrm{obj}+\mathcal{L}_\mathrm{rel} \label{con:eq_15}
	\end{equation}
\end{small}

Minimizing the total loss can improve the performance of both object classification and relation classification. This ensures that the generated scene graphs are more accurate.

\section{Experiments}
In this section, a series of experiments are conducted to demonstrate the effectiveness of TA-HDG. Sec.~\ref{4.1} gives a description of the benchmarks, evaluation metrics, and setups. Sec.~\ref{4.2} and ~\ref{4.3} analyze the experimental results on the Visual Genome and Open Images datasets, gradually demonstrating its capability in predicting scene graphs accurately and alleviating the long-tail distribution problem. Sec.~\ref{4.4} illustrates the prediction results on scene graphs.

\begin{table*}
	\centering
	\caption{Results on VG. The best result is bold. the second-best result is underlined.}
	\label{tab1}
	\begin{center}		
		\scalebox{1.25}	{
			\begin{tabular}{lrrrrrr}
				\multirow{2}{*}{ Model } & \multicolumn{2}{c}{ SGDet } & \multicolumn{2}{c}{ SGCls } & \multicolumn{2}{c}{ PredCls } \\
				\cmidrule(r){2-3}	\cmidrule(r){4-5}	\cmidrule(r){6-7}
				 & R@50/100 & mR@50/100 & R@50/100 & mR@50/100 & R@50/100 & mR@50/100 \\
				\hline\hline IMP\cite{xu2017scene} & 3.44 / 4.24 & - & 21.72 / 24.38 & - & 44.75 / 53.08 & - \\
				 Motif\cite{zellers2018neural} & 27.30 / 30.40 & - & 35.50 / 36.20 & - & 65.10 / 66.90 & - \\
				 CMAT\cite{chen2019counterfactual} & 27.90 / 31.20 & - & 39.00 / 39.80 & - & 66.40 / 68.10 & - \\
				 VCTree\cite{tang2019learning} & 27.90 / 31.30 & -/ 8.00 & 38.10 / 38.80 & -/ 10.80 & 66.40 / 68.10 & -/ 19.40 \\
				 HOSE-Net\cite{wei2020hose} & 28.90 / 33.30 & - & 36.30 / 37.40 & - & 66.70 / 69.20 & - \\
				 Part-Aware\cite{tian2020part} & 29.40 / 32.70 & 7.70 / 8.80 & 39.40 / 40.20 & 10.90 / 11.60 & 67.70 / 69.40 & 19.20 / 20.90 \\
				 GPS-Net\cite{lin2020gps} & 28.40 / 31.70 & -/ 9.80 & 39.20 / 40.10 & -/ 12.60 & 66.90 / 68.80 & -/ 22.80 \\
				 BGNN\cite{li2021bipartite} & 31.00 / 35.80 & 10.70 / 12.60 & 37.40 / 38.50 & 14.30 / 16.50 & 59.20 / 61.30 & 30.40 / 32.90 \\
				 HL-Net\cite{lin2022hl} & 33.70 / 38.10 & -/ 9.20 & \underline{42.60 / 43.50} & -/ 13.50 & 67.00 / 68.90 & -/ 22.80 \\
				 HetSGG\cite{yoon2023unbiased} & 30.00 / 34.60 & 12.20 / 14.40 & 37.60 / 38.70 & \underline{17.20 / 18.70} & 57.80 / 59.10 & \underline{31.60 / 33.50} \\
				 RU-Net\cite{lin2022ru} & 32.90 / 37.50 & -/ 10.80 & 42.40 / 43.30 & -/ 14.60 & \underline{67.70 / 69.60} & -/ 24.20 \\
				 NBP\cite{liu2023neural} & - & \underline{12.90 / 14.70} & - & 15.10 / 16.50 & - & 28.50 / 30.60 \\
				 CSL\cite{liu2023constrained} & - & 11.90 / 14.30 & - & 16.70 / 17.90 & - & 29.50 / 31.60 \\
				\hline TA-HDG & \textbf{33.71 / 38.21} & \textbf{14.85 / 16.66} & \textbf{43.26 / 44.74} & \textbf{22.02 / 23.13} & \textbf{67.78 / 69.98} & \textbf{32.34 / 34.19} \\
				
			\end{tabular}
		}
	\end{center}
\end{table*}
\subsection{Dataset, Metric, and Setup} \label{4.1}
\textbf{Visual Genome (VG).} The VG\cite{krishna2017visual} dataset comprises 108,249 images. The most frequent 150 object classes and 50 relation classes are employed for evaluation. It is the largest and most widely used dataset for the scene graph generation task. 

\textbf{Open Images (OI).} The OI \cite{kuznetsova2020open} dataset comprises 133,503 images, including 301 object classes and 31 relation classes. It is partitioned into 126,368 images for training, 1,813 for validation, and 5,322 for testing. 

\textbf{Metric.} The proposed model is evaluated on three standard subtasks: (1) Predicate Classification (\textbf{PredCls}), (2) Scene Graph Classification (\textbf{SGCls}), (3) Scene Graph Detection (\textbf{SGDet}). For the VG dataset, the model is evaluated via Recall (\textbf{R@K}) and mean Recall (\textbf{mR@K}). For the OI dataset, in addition to \textbf{R@K} and \textbf{mR@K}, the weighted mean AP of relations ($\mathbf{wmAP_{rel}}$), weighted mean AP of phrase ($\mathbf{wmAP_{phr}}$), and weighted metric score ($\mathbf{score_{wtd}}$) of the model are reported. Here, the $\rm {score_{wtd}}$ is calculated as: $\rm {score_{wtd}=0.2 \times R@50 + 0.4 \times wmAP_{rel}+0.4 \times wmAP_{phr}}$. In addition, to validate the performance of subject-object pair selection, the pair Recall (pR@K) is also computed by jointly considering the subject and object, temporarily ignoring the relation.

\textbf{Setup.} For a fair comparison, the object detector is a pre-trained Faster RCNN\cite{ren2016faster} with ResNeXt-101-FPN\cite{lin2017feature} as the backbone network. The layers before the ROIAlign layer are frozen. The total loss described in Sec.~\ref{3.3} is utilized to optimize the model on a NVIDIA TESLA V100 GPU with SGD. The initial learning rate, batch size, and weight decay are set to 0.008, 5, and 1.0e-05, respectively. The top-80 object proposals in each image are selected based on NSM per class, with an IoU of 0.5. In all experiments, the $s_\mathrm{b}$, $s_\mathrm{l}$, and $K$ are set to 600, 0.00001, and 4096, respectively.

\subsection{Performance on Visual Genome} \label{4.2}
In this section, the performance of the TA-HDG on VG is thoroughly evaluated. Sec.\ref{section:4.2.1} demonstrates the superiority of TA-HDG in all subtasks, particularly in alleviating the long-tail problem, by comparing it with various state-of-the-art methods. Sec.\ref{section:4.2.2} adds components incrementally to verify the importance of each module. Moreover, comparative experiments with different strategies confirm the superiority of the subject-object pair selection strategy. Finally, the efficacy of the relation categorization is evaluated. The experiment proves that this categorization can significantly adjust the distribution of relations and alleviate the long-tail problem.

\subsubsection{Comparisons with State-of-the-Art Methods}\label{section:4.2.1}
To evaluate the TA-HDG, it is compared with other state-of-the-art (SoTA) methods in four metrics (mR@50/100 and R@50/100) for three subtasks. The evaluative results are shown in Tab.~\ref{tab1}. The TA-HDG performs well on three subtasks: SGDet, SGCls, and PredCls. Particularly, it significantly outperforms HetSGG on mR@50 and mR@100 for SGCls by 4.82\% and 4.43\%, respectively. Specifically, compared with the SoTA method HL-Net, TA-HDG achieves improvements in mR@100 with 7.46\%, 9.63\%, and 11.39\% for the three tasks, respectively. It is speculated that HL-Net, which focuses on optimizing object features in ART, ignores the importance of relation features. This leads the model to favor head classes by data bias. Conversely, TA-HDG improves the performance of tail classes by categorizing relation types and performing message passing to refine relation features. Similarly, compared with HetSGG, a SoTA method, TA-HDG improves the performance by 3.61\%, 6.04\%, and 10.88\% on three tasks with R@100, respectively. It is argued that although HetSGG alleviates the long-tail problem by treating the scene graph as a heterogeneous graph, the decrease in R@K suggests that it may sacrifice the performance of head classes due to the neglect of the interactions between relations. Conversely, TA-HDG identifies head classes more accurately by combining heterogeneous and dual graphs, and selecting subject-object pairs with the distance, confidence, and existence information. In addition, although other methods, such as GPS-Net, BGNN, and RU-Net, improve either R@K or mR@K, they focus on a single metric. This may be because they ignore the semantic context among different objects with the same relation and among different relations with the same object. As a result, the neglect of intra-type context leads to the inability to fully understand and represent complex interactions, reducing the accurate prediction of head classes or tail classes. In contrast, TA-HDG excels in four metrics, proving its advantage in dealing with the unbalanced problem. 

\begin{figure*}[htbp]
	\centering
	\vspace{-0.3cm}
	\setlength{\belowcaptionskip}{-0.4cm}
	\includegraphics[width=182mm]{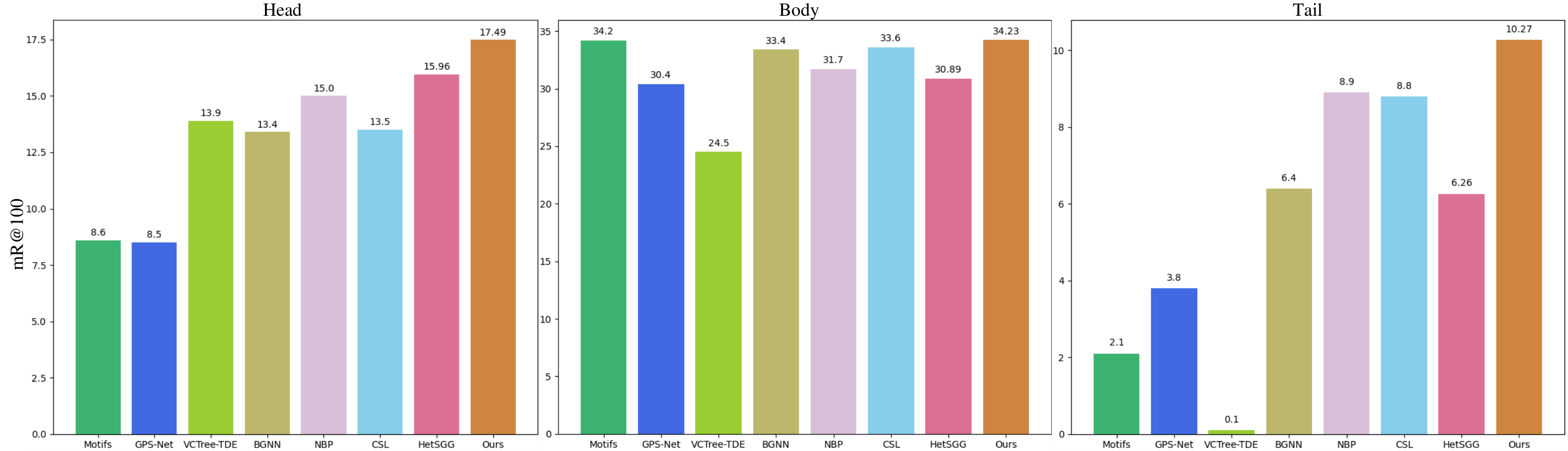}	
	\caption{Results of mR@100 on the head, body, and tail classes in SGDet task. The $x$-axis is various methods, the $y$-axis is mR@100. The left is the head performance, the middle is the body performance, the right is the tail performance.
	}
	\label{comparation}	
\end{figure*}

To verify its ability to handle long-tail relation distributions, the mR@K of different methods under each relation distribution are compared. The comparative results in SGDet are shown in Fig.~\ref{comparation}. Experiments indicate that TA-HDG outperforms the HetSGG by a 1.53\% margin in body and NBP by a 1.37\% margin in Tail. The mR@K of TA-HDG on the body and tail is significantly higher than other methods by approximately 4\%. This is first attributed to the distribution of tail classes adjusted in TA-HDG by categorizing relation types. Second, the model facilitates the understanding of tail classes by capturing intra- and inter-type context for each relation type. In head, TA-HDG reduces more meaningless edges in heterogeneous and dual graphs by introducing distance, confidence, and existence information. Thus, TA-HDG shows superior performance compared with other methods. These results confirm that TA-HDG not only optimizes the performance of tail classes, but also improves the accuracy of head classes. This further confirms that TA-HDG considers the performance of each relation.

\begin{table}
	\centering
	\caption{Results of different modules in SGDet task. The best result is bold. The \textbf{A} represents the Inital Graph. The \textbf{B} represents the Heterogeneous Graph \& Inter-Type MP. The \textbf{C} represents the Dual Graph \& Intra-Type MP.}
	\label{tab2}
	\begin{center}		
		\scalebox{0.98}{
			\begin{tabular}{cccccccc}
				\multirow{2}{*}{ No. } & \multicolumn{3}{c}{ Method } & \multirow{2}{*}{R@50} & \multirow{2}{*}{ R@100 } & \multirow{2}{*}{mR@50} & \multirow{2}{*}{ mR@100 } \\
				\cline{2-4} & \textbf{A} & \textbf{B} & \textbf{C} & & & & \\
				\hline \hline1 & \ding{53} & \ding{53} & \ding{53} & 29.27 & 33.83 & 11.18 & 12.87 \\
				2 & \ding{53} & \ding{53} & \checkmark & 31.22 & 35.91 & 11.87 & 13.91 \\
				3 & \ding{53} & \checkmark & \ding{53} & 30.95 & 34.89 & 13.29 & 15.31 \\
				4 & \ding{53} & \checkmark& \checkmark & 32.45 & 36.59 & 14.16 & 15.82 \\
				5 & \checkmark & \ding{53} & \ding{53} & 30.58 & 35.64 & 11.48 & 13.49 \\
				6 & \checkmark & \ding{53} & \checkmark & 32.54 & 37.69 & 12.27 & 14.27 \\
				7 & \checkmark & \checkmark & \ding{53} & 32.21 & 36.84 & 13.64 & 15.83 \\
				8 & \checkmark & \checkmark & \checkmark & \textbf{33.71} & \textbf{38.21} & \textbf{14.85} & \textbf{16.66} \\
				\hline
			\end{tabular}
		}
	\end{center}
\end{table}
\begin{figure}[htbp]
	\centering
	\vspace{-0.3cm}
	\setlength{\belowcaptionskip}{-0.4cm}
	\includegraphics[width=87mm]{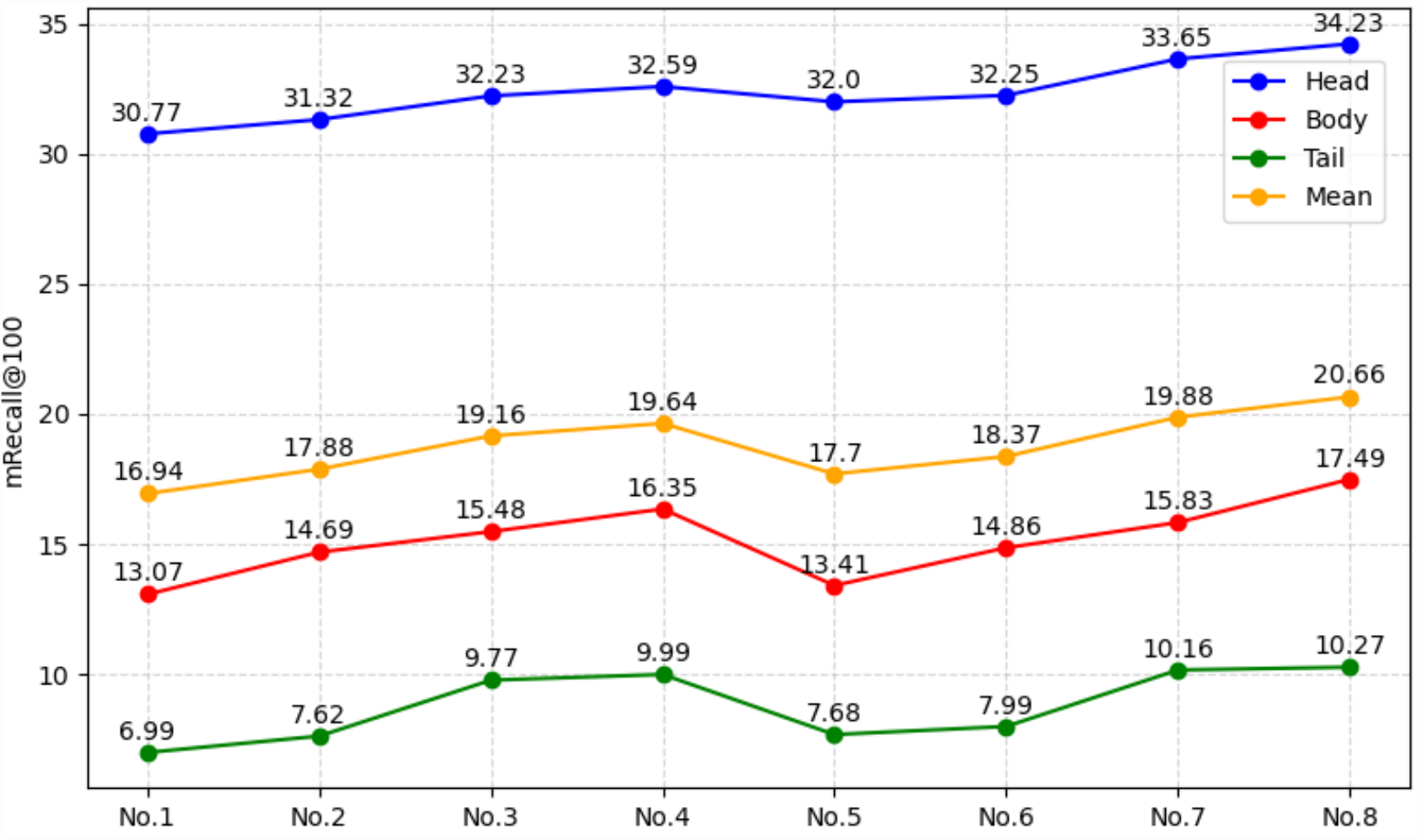}
	\caption{Results of each long-tail group under different modules in SGDet with mR@100. The model numbers correspond to Table~\ref{tab2}.
	}
	\label{Ablation}	
\end{figure}
\subsubsection{Ablation Studies}\label{section:4.2.2}
\textbf{Effectiveness of the Proposed Modules.} An ablation study is performed to verify the effectiveness of HDGC and TAMP. To ensure fairness in comparison, components are added progressively to the baseline and all experimental settings remain the same as TA-HDG. Tab.~\ref{tab2} shows the evaluative results under different combinations of components on R@K and mR@K for SGDet. From No.1 to No.8, the performance shows continuous improvement as more components are integrated. Specifically, according to the comparison between No.1 and No.4, TAMP improves the four metrics by approximately 3 percent. This proves that the TAMP makes a significant contribution to improving the overall performance and alleviating the long-tail problem. This is because the TAMP synthesizes the intra- and inter-type context to understand the complex interactions within a scene. Comparing No.4 with No.8, the removal of the Initial Graph, i.e., the subject-object pair selection strategy, results in a 1\% decrease in four metrics. This suggests that selecting meaningful subject-object pairs utilizing rich information of objects is important for modeling the interactions. Comparing No.6 with No.8, the removal of the Heterogeneous Graph \& Inter-Type MP decreases by at least 2\% in mR@K. This proves the key role of two components in alleviating the long-tail problem, that is, adjusting unbalanced distribution by relations categorization and understanding inter-type interactions by inter-type context aggregation. By comparing No.7 with No.8, removing the Dual Graph \& Intra-Type MP results in a 1.5\% significant decrease of R@K. This proves the importance of modeling and capturing the intra-type interactions in refine features to improve the accurate prediction of relations. In summary, each component plays a pivotal role in improving the performance, and TA-HDG utilizes the synergy among them to achieve significant performance improvement. 

To further investigate the impact of different components in the long-tail problem, the mR@K of components under long-tail distribution conditions are evaluated, as shown in Fig.~\ref{Ablation}. The trends of No.1-4 and No.5-8 in the line graph indicate that the performance of each distribution condition increases linearly, as the number of modules in TA-HDG increases linearly. Specifically, comparing No.1 with No.5 reveals that introducing the subject-object pair selection strategy achieves performance improvement for three different distributions. This indicates that the combination of distance, confidence, and existence information in selecting subject-object pairs causes a positive impact on the long-tail problem. Further comparing No.6 with No.8 shows that after removing the Heterogeneous Graph \& Inter-Type MP, the decrease of performance in No.6 on body and tail significantly exceeds that on head. Specifically, when it is difficult to distinguish between interactive and non-interactive relation types, and it is difficult to model and capture the interactions between objects and relations, the performance of No.6 decreases by 1.98\% on head, 2.63\% on body, and 2.28\% on tail compared with that of No.8. This finding proves that modeling the interactions between objects and performing inter-type message passing, not only maintains the performance of head, but also optimizes the performance of body and tail. Similarly, No.7 and No.8 also show the same variation trend, which indicates that modeling the interactions between relations, as well as capturing the intra-type context, can also effectively improve the prediction performance of relations. Moreover, the trend of No.1-8 shows that although the influence of TA-HDG on head is not as significant as on body or tail, the performance of head still maintains a steady upward trend. This proves the critical role of each module in improving the recognition of head and tail classes.

\begin{table}
	\centering
	\caption{Results of different subject-object pair selection strategy in SGDet task. The best result is bold.}
	\label{tab3}
	\begin{center}		
		\scalebox{1.08}{
			\begin{tabular}{lcccc}
				\multirow{2}{*}{ Method } & \multicolumn{2}{c}{ Visual Phrases } & \multicolumn{2}{c}{ Object-Pair Proposals} \\
				\cmidrule(r) { 2 - 3 }\cmidrule(r) { 4-5 } & R@50 & R@100 & pR@50 & pR@100 \\
				\hline \hline $\mathrm{{TA\mbox {-}HDG}_{con}}$ & 32.28 & 36.01 & 43.70 & 50.35 \\
				$\mathrm{{TA\mbox {-}HDG}_{IoU}}$ & 33.04 & 35.94 & 43.25 & 49.75 \\
				$\mathrm{{TA\mbox {-}HDG}_{IoU}^+}$ & 32.92 & 36.70 & 44.68 & 51.59 \\
				$\mathrm{{TA\mbox {-}HDG}_{sim}}$ & 32.16 & 36.15 & 43.99 & 51.09 \\
				$\mathrm{{TA\mbox {-}HDG}_{dis}}$ & 32.31 & 36.29 & 45.02 & 52.34 \\
				$\mathrm{{TA\mbox {-}HDG}_{lin}}$ & 33.11 & 38.14 & 45.18 & 52.46 \\
				$\mathrm{{TA\mbox {-}HDG}_{dis+sim}}$ & 32.95 & 37.94 & 45.13 & 52.42 \\
				$\mathrm{{TA\mbox {-}HDG}_{con+lin}}$ & 33.43 & 38.19 & 45.35 & 52.61 \\
				$\mathrm{{TA\mbox {-}HDG}_{dis+lin}}$ & 33.38 & 38.11 & 45.30 & 52.53 \\
				$\mathrm{{TA\mbox {-}HDG}}$ & \textbf{33.71} & \textbf{38.21} & \textbf{45.36} & \textbf{52.62} \\
				\hline
			\end{tabular}
		}
	\end{center}
\end{table}

\textbf{Design Choices in Subject-Object Pair Selection Strategy.} Ablation studies are conducted on different strategies to determine an effective strategy for subject-object pair selection. Tab.~\ref{tab3} summarizes the results of these strategies. Among them in selecting pairs, the $\mathrm{{TA\mbox {-}HDG}_{con}}$ bases on confidence alone, the $\mathrm{{TA\mbox {-}HDG}_{IoU}}$ calculates the IoU of subject-object pairs, the IoU in $\mathrm{{TA\mbox {-}HDG}_{IoU}^+}$ is divided by the area of the smaller bounding box between the subject and the object, the $\mathrm{{TA\mbox {-}HDG}_{sim}}$ measures the semantic similarity of subject-object pairs by the cosine similarity, the $\mathrm{{TA\mbox {-}HDG}_{dis}}$ calculates the center distance between subject-object pairs, the $\mathrm{{TA\mbox {-}HDG}_{lin}}$ looks up the existence probability of subject-object pairs based on the co-linearity, the $\mathrm{{TA\mbox {-}HDG}_{dis+sim}}$ combines the $\mathrm{{TA\mbox {-}HDG}_{dis}}$ and the $\mathrm{{TA\mbox {-}HDG}_{sim}}$, the $\mathrm{{TA\mbox {-}HDG}_{con+lin}}$ combines the $\mathrm{{TA\mbox {-}HDG}_{con}}$ and the $\mathrm{{TA\mbox {-}HDG}_{lin}}$, the $\mathrm{{TA\mbox {-}HDG}_{dis+lin}}$ combines the $\mathrm{{TA\mbox {-}HDG}_{dis}}$ and the $\mathrm{{TA\mbox {-}HDG}_{lin}}$.

\begin{figure}[htbp]
	\centering
	\vspace{-0.3cm}
	\setlength{\belowcaptionskip}{-0.4cm}
	\includegraphics[width=89mm]{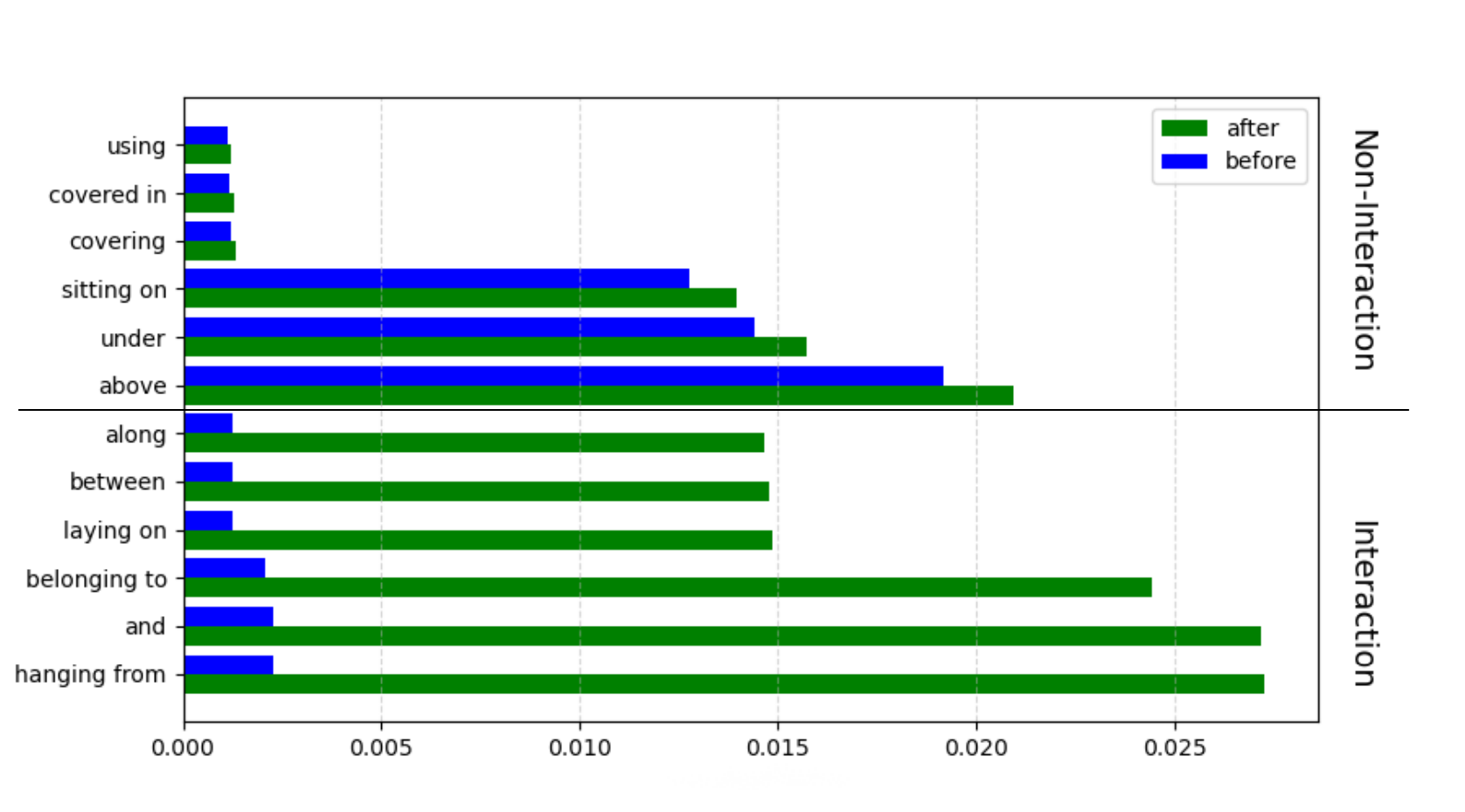}
	\caption{Comparison of distribution ratios before and after relation categorization on VG. The $x$-axis represents the distribution ratios of relations, the $y$-axis represents the relation types.
	}
	\label{vg}	
\end{figure}

These results show that TA-HDG performs best in terms of visual phrases (R@K) and subject-object pair proposals (pR@K). Specifically, by comparing with the traditional method $\mathrm{{TA\mbox {-}HDG}_{con}}$, TA-HDG achieves an improvement of about 2\% in R@100 and pR@100. This indicates that the method effectively reduces the incorrect selection of subject-object pairs by introducing distance, confidence, and existence information. Further observing $\mathrm{{TA\mbox {-}HDG}_{IoU}}$ and $\mathrm{{TA\mbox {-}HDG}_{IoU}^+}$, $\mathrm{{TA\mbox {-}HDG}_{IoU}}$ is lower than $\mathrm{{TA\mbox {-}HDG}_{IoU}^+}$ by 1.84\% on pR@100. This may be because $\mathrm{{TA\mbox {-}HDG}_{IoU}^+}$ increases the sensitivity to small objects by normalizing the IoU by the area of the smaller object. This helps to improve the accurate recognition of the relation between small and large objects. In addition, $\mathrm{{TA\mbox {-}HDG}_{sim}}$ is lower than $\mathrm{{TA\mbox {-}HDG}_{dis+sim}}$ by 1.79\% on R@100. Similarly, $\mathrm{{TA\mbox {-}HDG}_{sim}}$ is lower than TA-HDGdis+sim by 1.65\%. These indicate that selecting subject-object pairs only based on semantic or spatial information will bring error propagation. Therefore, a combination of both is needed. In summary, these results confirm the effectiveness of TA-HDG, which significantly improves the accuracy of subject-object pair proposals and reduces meaningless edges through rich information of objects.

\textbf{Effectiveness of the relation categorization.} To analyze the effectiveness of the relation categorization, Fig.~\ref{vg} shows the changes in the distribution ratios of relations before and after the categorization. After the categorization, there is a significant increase in the distribution ratios of relations, particularly in the interactive relation type. For instance, the relation \textit{hanging from} is only 0.0023 for the distribution ratio in the entire dataset, but after the categorization, its ratio in the interactive relation type increases to 0.0272. This significant increase proves the effectiveness of the relation categorization. Furthermore, improving the distribution of each relation reduces the impact of data bias on the model. This allows the model to better learn from less frequent but important tail classes, helping alleviate the long-tail problem.

\subsection{Performance on Open Images} \label{4.3}
This section systematically evaluated the performance on OI to verify its generalization and effectiveness. Sec.\ref{section:4.3.1} shows that TA-HDG significantly advantages over the SoTA methods on all evaluation metrics through detailed comparative analyses. This further proves that TA-HDG is effective and generalizable when dealing with different datasets. Sec.\ref{section:4.3.2} also employs an incremental method to prove the key role of each module in improving model performance. Furthermore, the subject-object pair selection strategy applied on VG is transferred to OI, thereby verifying its generalization. Finally, the effect of relation categorization is evaluated and its results is in line with the results on VG.

\subsubsection{Comparisons with State-of-the-Art Methods}\label{section:4.3.1}
To verify the effectiveness and generalization of TA-HDG in dealing with various datasets, it is evaluated on OI for SGDet. Tab.~\ref{tab4} shows the comparative results. All methods utilize a uniform object detector to ensure a fair comparison. On this basis, TA-HDG achieves significant improvements in all metrics. This proves the generalization of TA-HDG. Specifically, owing to the utilization of the $\ell_{\mathrm{p}}$-based graph regularization to filter subject-object pairs, the performance of RU-Net in $\rm{wmAP_{rel}}$, which is 35.40\%, is comparable to TA-HDG with 35.67\%. However, RU-Net is weaker than TA-HDG in terms of R@50 and $\rm{wmAP_{phr}}$, with 76.9\% and 34.9\%, respectively. In contrast, TA-HDG maintains high performance in these metrics, especially on the comprehensive metric $\rm{score_{wtd}}$, which achieves 45.19\%. These data prove that selecting meaningful subject-object pairs by introducing distance, confidence, and existence information improves the quality of scene graphs. Furthermore, compared with HetSGG, which performs best in mR@50, TA-HDG achieves an improvement in R@50 with 4.91\%. This result further indicates that HetSGG only improves the performance of tail classes, while TA-HDG simultaneously improves the performance of head classes and tail classes by modeling and capturing the intra- and inter-type context for interactive and non-interactive relation types. In conclusion, TA-HDG maintains stable performance in dealing with both common relations and rare relations through subject-object pair selection strategy and Type-Aware message passing.

\begin{table}
	\centering
	\caption{Results on OI. The best result is bold. The second-best result is underlined.}
	\label{tab4}
	\begin{center}		
		\scalebox{0.93}{
			\begin{tabular}{lccccc}
				 Model & mR@50 & R@50 & $\mathrm{wmAP_{rel}}$ & $\mathrm{wmAP_{phr}}$  &  $\mathrm{score_{wtd}}$ \\
				\hline \hline GPS-Net\cite{lin2020gps} & 38.90 & 74.70 & 32.80 & 33.90 & 41.60 \\
				 BGNN\cite{li2021bipartite} & 40.45 & 74.95 & 33.51 & 34.15 & 42.06 \\
				 HL-Net\cite{lin2022hl} & - & 76.50 & 35.10 & 34.70 & 43.20 \\
				 HetSGG\cite{yoon2023unbiased} & \underline{42.70} & 76.80 & 34.60 & 35.50 & 43.30 \\
				 RU-Net\cite{lin2022ru} & - & \underline{76.90} & \underline{35.40} & 34.90 & \underline{43.50} \\
				 NBP\cite{liu2023neural} & 41.97 & 75.54 & 34.44 & \underline{35.66} & 43.08 \\
				 CSL\cite{liu2023constrained} & 41.72 & 75.44 & 34.30 & 35.38 & 42.86 \\
				\hline TA-HDG & \textbf{43.28} & \textbf{81.71} & \textbf{35.67} & \textbf{36.46} & \textbf{45.19} \\
				
			\end{tabular}
		}
	\end{center}
\end{table}

\subsubsection{Ablation Studies}\label{section:4.3.2}
\textbf{Effectiveness of the Proposed Modules.} To verify the effectiveness of each component on OI, the components are progressively added to the general baseline model. The strategy and settings are the same as VG. The results are summarized in Tab.~\ref{tab5}. Experiments No.1 to No.8 demonstrate the significant role of each module in enhancing model performance. Specifically, by comparing the traditional message passing in No.1 and the intra-type message passing in No.2, TA-HDG shows the significant performance improvement in all metrics by adding the Dual Graph \& Intra-Type MP, especially in R@50 with at least 1\%. This suggests that capturing the inter-type context and modeling the interactions between objects, without considering the intra-type context and the interactions between relations, may limit the understanding of complex interactions, reducing predictive accuracy for relations. Further comparing No.1 with No.3, when the Heterogeneous Graph \& Inter-Type MP is added, the improvement on mR@50 exceeds other metrics. This proves the importance of categorizing relation types and employing inter-type message passing for these relations types in alleviating the long-tail problem. By comparing No.1 and No.5, the composite metric, $\rm {score_{wtd}}$, rises by 0.4\% after introducing the Initial Graph. Here, No.1 selects subject-object pairs based on confidence, while No.5 employs spatial, semantic, and co-linearity information. This result proves that the traditional subject-object pair selection method leads to more meaningless edges an d incorrect subject-object pairs, affecting the overall performance of the model. By comparing No.5 with No.8, the model degenerates to an early method that relies on local information for predicting relations if removing the TAMP. Meanwhile, the performance on each metric drops by more than 1\%. This change confirms that context significantly improves the performance of relation prediction.

\begin{table}
	\centering
	\caption{Results of different modules in SGDet task. The best result is bold. The \textbf{A} represents the Inital Graph. The \textbf{B} represents the Heterogeneous Graph \& Inter-Type MP. The \textbf{C} represents the Dual Graph \& Intra-Type MP.}
	\label{tab5}
	\begin{center}		
			\scalebox{0.83}{
			\begin{tabular}{ccccccccc}
				 \multirow{2}{*}{ No. } & \multicolumn{3}{c}{ Method } & \multirow{2}{*}{mR@50} & \multirow{2}{*}{R@50} & \multirow{2}{*}{$\mathrm{wmAP_{rel}}$} & \multirow{2}{*}{$\mathrm{wmAP_{pht}}$} & \multirow{2}{*}{$\mathrm{score_{wtd}}$} \\
				\cline{2-4} & \textbf{A} & \textbf{B} & \textbf{C} & & & & & \\
				\hline \hline1 & \ding{53} & \ding{53} & \ding{53} & 41.95 & 78.62 & 33.95 & 35.19 & 43.38 \\
				 2 & \ding{53} & \ding{53} &\checkmark & 42.17 & 79.98 & 34.53 & 35.50 & 44.01 \\
				 3 & \ding{53} &\checkmark & \ding{53} & 42.76 & 79.33 & 34.47 & 35.68 & 43.93 \\
				 4 & \ding{53} &\checkmark &\checkmark & 42.98 & 80.46 & 35.04 & 35.97 & 44.50 \\
				 5 &\checkmark & \ding{53} & \ding{53} & 42.04 & 79.19 & 34.32 & 35.43 & 43.74 \\
				 6 &\checkmark & \ding{53} &\checkmark & 42.26 & 80.55 & 34.90 & 35.74 & 44.37 \\
				 7 &\checkmark &\checkmark & \ding{53} & 42.85 & 79.90 & 34.84 & 35.92 & 44.28 \\
				 8 &\checkmark &\checkmark &\checkmark & \textbf{43.28} & \textbf{81.71} & \textbf{35.67} & \textbf{36.46} & \textbf{45.19} \\
				\hline
			\end{tabular}
		}
	\end{center}
\end{table}

\begin{table}
	\centering
	\caption{Results of different subject-object pair selection strategy in SGDet task. The best result is bold.}
	\label{tab6}
	\begin{center}		
		\scalebox{0.85}{
			\begin{tabular}{lccccc}
				Method & mR@50 & R@50 & $\mathrm{wmAP_{rel}}$ & $\mathrm{wmAP_{phr}}$ & $\mathrm{score_{wtd}}$ \\
				\hline\hline $\mathrm{{TA\mbox {-}HDG}_{con}}$ & 42.94 & 80.31 & 34.27 & 36.28 & 44.28 \\
				$\mathrm{{TA\mbox {-}HDG}_{IoU}}$ & 41.91 & 80.16 & 33.61 & 34.93 & 43.45 \\
				$\mathrm{{TA\mbox {-}HDG}_{IoU}^+}$ & 42.6 & 80.92 & 34.16 & 35.43 & 44.02 \\
				$\mathrm{{TA\mbox {-}HDG}_{sim}}$ & 41.17 & 81.04 & 33.40 & 33.59 & 43.00 \\
				$\mathrm{{TA\mbox {-}HDG}_{dis}}$ & 41.62 & 80.28 & 33.47 & 34.19 & 43.12 \\
				$\mathrm{{TA\mbox {-}HDG}_{lin}}$ & 43.1 & 81.11 & 35.60 & 36.30 & 44.98 \\
				$\mathrm{{TA\mbox {-}HDG}_{dis+sim}}$ & 42.46 & 81.29 & 34.06 & 34.25 & 43.58 \\
				$\mathrm{{TA\mbox {-}HDG}_{con+lin}}$ & 43.22 & 81.38 & 35.57 & 36.37 & 45.05 \\
				$\mathrm{{TA\mbox {-}HDG}_{dis+lin}}$ & 43.27 & 81.43 & 35.65 & 36.45 & 45.13 \\
				$\mathrm{{TA\mbox {-}HDG}}$ & \textbf{43.28} & \textbf{81.71} & \textbf{35.67} & \textbf{36.46} & \textbf{45.19} \\
				\hline
			\end{tabular}
		}
	\end{center}
\end{table}
\textbf{Design Choices in Subject-Object Pair Selection Strategy.} A series of different subject-object pair selection strategies are compared to verify the generalization of the subject-object pair selection strategy applied by TA-HDG. These strategies remain the same as those of VG. Tab.~\ref{tab6} shows the comparison results of various strategies. Although the performance of various schemes varies in two datasets, TA-HDG shows the best performance on both datasets. Specifically, regarding the $\rm{score_{wtd}}$ on OI, the second-performing method is $\mathrm{{TA\mbox {-}HDG}_{dis+lin}}$, and the worst-performing method is $\mathrm{{TA\mbox {-}HDG}_{sim}}$. For R@100 on VG, $\mathrm{{TA\mbox {-}HDG}_{con+lin}}$ and $\mathrm{{TA\mbox {-}HDG}_{IoU}}$ rank as the second-performing and worst-performing, respectively. Other metrics can draw similar conclusions. These discrepancies may be because of the differences between the two datasets. Nonetheless, TA-HDG, which performs best on VG, maintains its best performance on OI. In summary, the subject-object pair selection strategy applied on VG can be effectively transferred to OI, while maintaining its superior performance. This proves the generalization of this strategy.

\begin{figure}[htbp]
	\centering
	\vspace{-0.3cm}
	\setlength{\belowcaptionskip}{-0.4cm}
	\includegraphics[width=90mm,height=40mm]{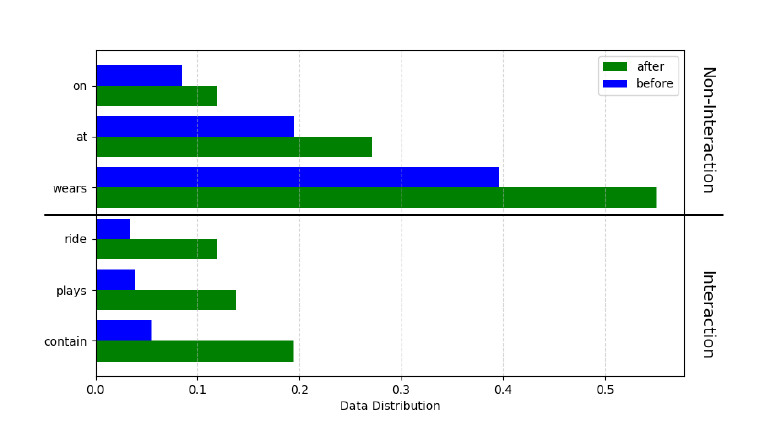}
	\caption{Comparison of distribution before and after relation categorization on OI. The $x$-axis represents the distribution of relations, the $y$-axis represents the relation types.
	}
	\label{oi}	
\end{figure}
\textbf{Effectiveness of the relation categorization.} To analyze the applicability of the categorization of interactive and non-interactive relations on OI, Fig.~\ref{oi} shows the distribution ratios of relations before and after the categorization. From this figure, a similar conclusion for OI can be drawn by observation, that is, the distribution ratio of each relation is significantly improved after the categorization. This proves the effectiveness and generalization of this categorization. In the case of the relation \textit{contain}, its distribution ratio is only 5.40\% in the entire dataset, but after the categorization, the distribution ratio increases to 19.40\% in the interactive type. This change indicates that categorizing relation classes into interactive and non-interactive relation types can significantly improve the distribution of various relations. This increases the diversity of tail classes, thereby alleviating the long-tail problem.

\subsection{Qualitative Analysis} \label{4.4}
To better understand the abilities of TA-HDG in accurately predicting scene graphs as well as alleviating the long-tail distribution problem, the predicted scene graphs by HetSGG and TA-HDG are compared. Among them, HetSGG employs traditional methods to select subject-object pairs and pass messages. Specifically, it selects subject-object pairs based on confidence while only focusing on the inter-type messages. 

\begin{figure}[htbp]
	\centering
	\vspace{-0.3cm}
	\setlength{\belowcaptionskip}{-0.4cm}
	\includegraphics[width=89mm]{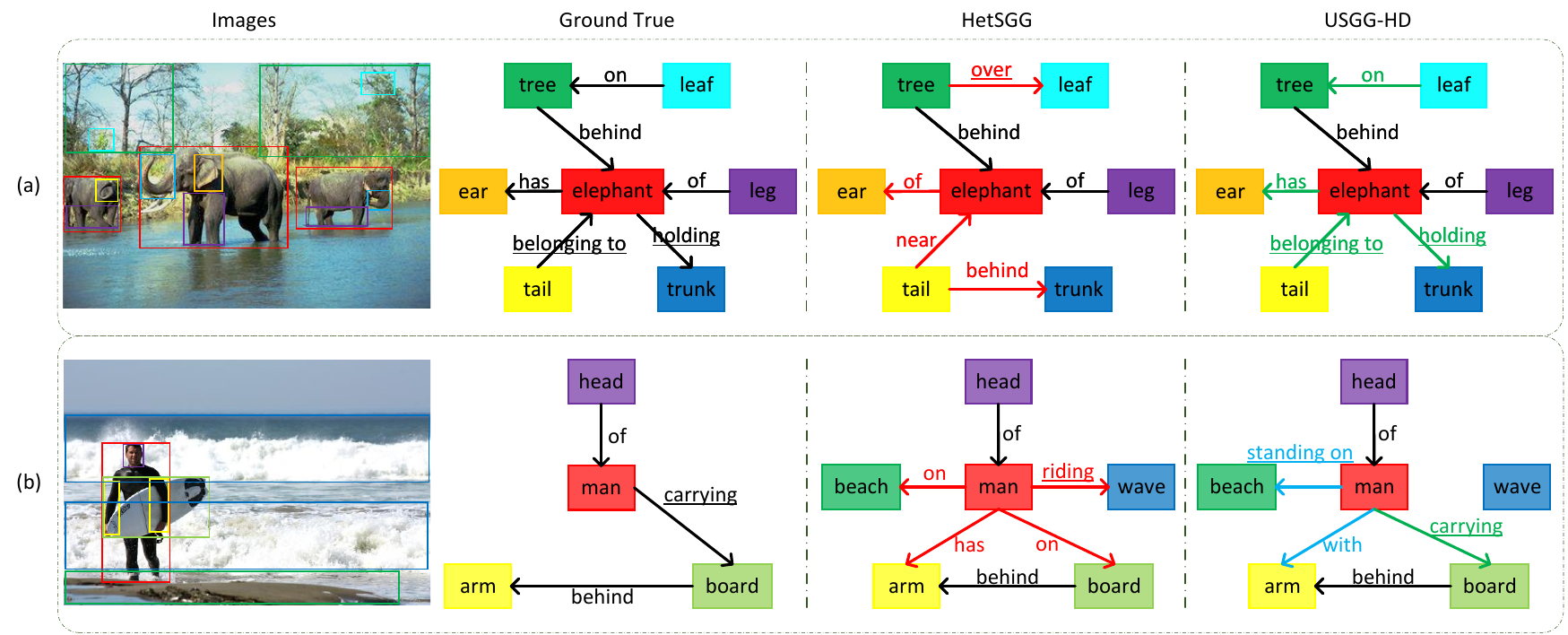}
	\caption{Scene graphs of TA-HDG and HetSGG on VG. (Red: incorrect predictions by HetSGG. Green: correct predictions by TA-HDG, but incorrect predictions by HetSGG. Blue: incorrect predictions by TA-HDG. The relations in the Body and Tail classes are underlined.)
	}
	\label{example1}
	
\end{figure}

The visualizations of VG are shown in Fig.~\ref{example1}. In Fig.~\ref{example1} (a), HetSGG fails to identify the \textit{elephant-trunk} pair. However, owing to combining distance, confidence, and existence information, TA-HDG detects the relation between \textit{elephant} and \textit{trunk}, thus enhancing the precision in subject-object pair selection. In Fig.~\ref{example1} (b), HetSGG incorrectly selects the \textit{man-wave} pair. However, owing to the comprehensive information being utilized to reduce meaningless edges and retain meaningful subject-object pairs, TA-HDG identifies them as meaningless, thus alleviating the error propagation. In addition, the statistical analysis on HetSGG reveals that such errors constitute up to 30\%. This confirms the limitation of the dependence on confidence alone for selecting subject-object pairs. In contrast, TA-HDG effectively enhances the precision in subject-object pair selection by considering distance, confidence, and existence information. 

\begin{figure}[htbp]
	\centering
	\vspace{-0.3cm}
	\setlength{\belowcaptionskip}{-0.4cm}
	\includegraphics[width=89mm]{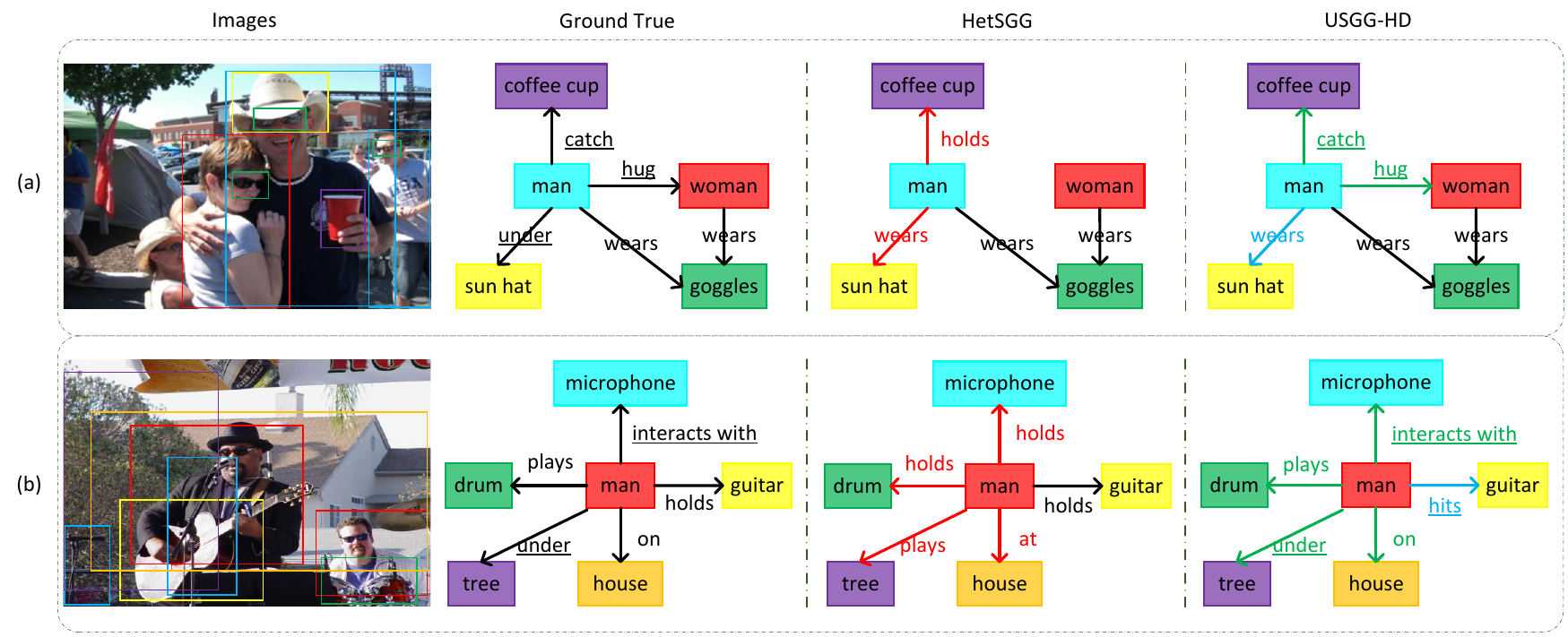}
	\caption{Scene graphs of TA-HDG and HetSGG on OI. (Red: incorrect predictions by HetSGG. Green: correct predictions by TA-HDG, but incorrect predictions by HetSGG. Blue: incorrect predictions by TA-HDG. The relations in the Body and Tail classes are underlined.)
	}
	\label{example2}
	
\end{figure}

The visualizations of OI are shown in Fig.~\ref{example2}. In Fig.~\ref{example2} (a), the relation between \textit{man} and \textit{coffee cup} in HetSGG, which is influenced by data bias, is incorrect (\textit{holds}). However, TA-HDG accurately predicts the rare relation (\textit{catch}) by adjusting the distribution of relations, as well as modeling and capturing the intra- and inter-type context of interactive and non-interactive relation types. Similarly, in Fig.~\ref{example2} (b), HetSGG predicts the relation between \textit{man} and \textit{microphone} as the head relation (\textit{holds}), whereas TA-HDG predicts the tail relation (\textit{interacts with}). Consequently, compared with HetSGG, TA-HDG exhibits significant improvement on body and tail classes (relations underlined). This indicates that categorizing the relation types and modeling and capturing the intra- and inter-type context can reduce the mispredictions caused by the long-tail data distribution.

\section{Conclusion}
This study proposes an unbiased SGG framework, called TA-HDG, that can address several challenges in SGG. The combination of heterogeneous and dual graphs overcoming the issue of simultaneously modeling the interactions between objects and the interactions between relations. The subject-object pair selection strategy reduces meaningless edges in heterogeneous and dual graphs, improving the accuracy of subject-object pairs. In addition, the TAMP addresses the limitation of capturing the semantic context among different objects with the same relations and among different relations with the same object, improving the accurate prediction of relations. The experimental results demonstrate the superiority of TA-HDG. However, owing to the fluctuating performance of pre-trained object detectors, which serve detected objects as nodes in scene graphs, TA-HDG struggles with generalization for various datasets. Meanwhile, the high computational cost caused by constructing dual graphs affects the efficiency. In the future, we will focus on Panoptic Scene Graph Generation, which is a method independent of object detectors to improve the independence of the model. In addition, we will explore efficient algorithms to construct dual graphs to reduce the computational cost.

\bibliographystyle{IEEEtran}
\bibliography{references.bib}   
\makeatletter

\end{document}